\documentclass[11pt,a4paper]{article}
\usepackage[hyperref]{emnlp-ijcnlp-2019}
\usepackage{times}
\usepackage{latexsym}

\usepackage{url}

\aclfinalcopy %

\usepackage{hyperref}
\usepackage{url}
\usepackage{subcaption} 
\usepackage{booktabs} %
\usepackage{enumitem}
\usepackage{graphicx}
\usepackage{bbm}
\usepackage{amsmath,amsfonts,bm}
\usepackage[T1]{fontenc}
\usepackage[utf8]{inputenc}

\usepackage{algorithm}
\usepackage{algpseudocode}

\title{Leveraging Just a Few Keywords for Fine-Grained Aspect Detection\\
Through Weakly Supervised Co-Training}

\author{Giannis Karamanolakis, Daniel Hsu, Luis Gravano\\
Columbia University, New York, NY 10027, USA \\
\texttt{\{gkaraman, djhsu, gravano\}@cs.columbia.edu}
}

\date{}

\begin{document}
\maketitle
\begin{abstract}
User-generated reviews can be decomposed into fine-grained segments (e.g., sentences, clauses), each evaluating a different aspect of the principal entity (e.g., price, quality, appearance). 
Automatically detecting these aspects can be useful for both users and downstream opinion mining applications. 
Current supervised approaches for learning aspect classifiers require many fine-grained aspect labels, which are labor-intensive to obtain. 
And, unfortunately, unsupervised topic models often fail to capture the aspects of interest. In this work, we consider weakly supervised approaches for training aspect classifiers that only require the user to provide a small set of seed words (i.e., weakly positive indicators) for the aspects of interest. First, we show that current weakly supervised approaches do not effectively leverage the predictive power of seed words for aspect detection. Next, we propose a student-teacher approach that effectively leverages seed words in a bag-of-words classifier (teacher); in turn, we use the teacher to train a second model (student) that is potentially more powerful (e.g., a neural network that uses pre-trained word embeddings). Finally, we show that iterative co-training can be used to cope with noisy seed words, leading
to both improved teacher and student models. Our proposed approach consistently outperforms previous weakly supervised approaches (by 14.1 absolute F1 points on average) in six different domains of product reviews and six multilingual datasets of restaurant reviews.
\end{abstract}

\section{Introduction}
\begin{figure}[t]
\centering
\includegraphics[scale=0.45]{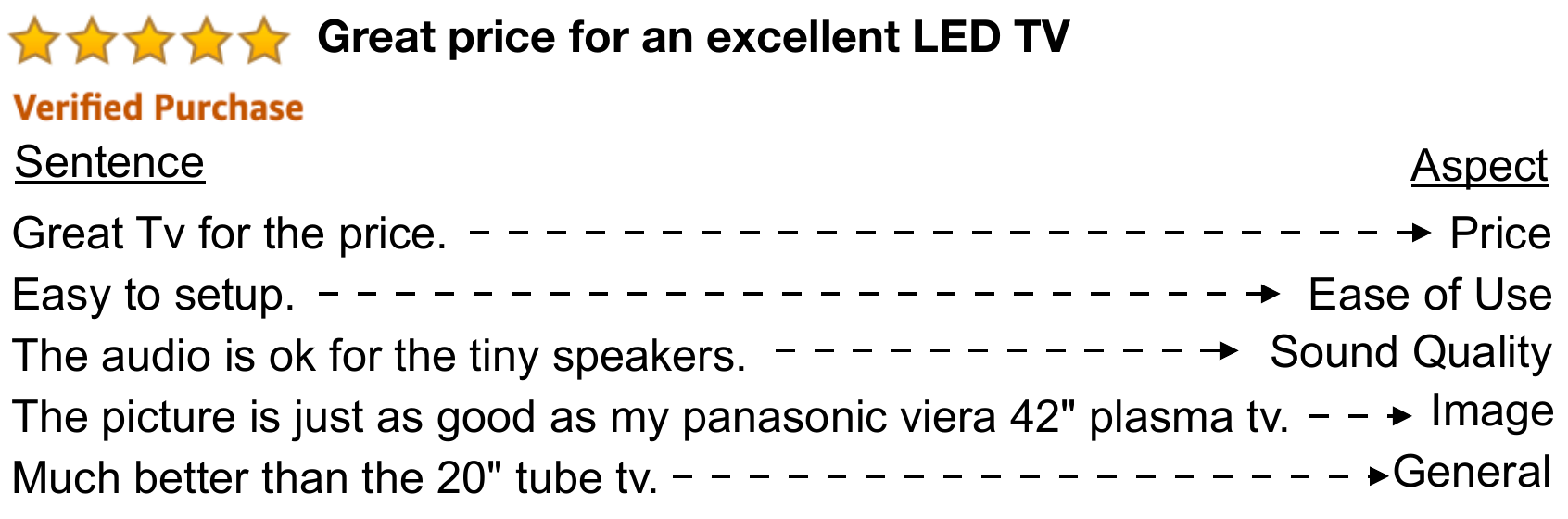}
\caption{Example of product review with aspect annotations: each individual sentence of the review discusses a different aspect (e.g., price) of the TV.}
\label{fig:amazon_tv_example}
\end{figure}

A typical review of an entity on platforms such as Yelp and
Amazon discusses multiple aspects of the entity (e.g., price, quality) in individual review segments (e.g., sentences, clauses).
Consider for example the Amazon product review in Figure~\ref{fig:amazon_tv_example}.
The text discusses various aspects of the TV such as price, ease of use, and sound quality.
Given the vast number of online reviews, both sellers and customers would benefit from automatic methods for detecting fine-grained segments that discuss particular aspects of interest. 
Fine-grained aspect detection is also a key task in downstream applications such as aspect-based sentiment analysis and multi-document summarization~\citep{hu2004mining,liu2012sentiment,pontiki2016semeval,angelidis2018summarizing}.

In this work, we consider the problem of classifying individual segments of reviews to pre-defined aspect classes when ground truth aspect labels are not available.
Indeed, reviews are often entered as unstructured, free-form text and do not come with aspect labels.
Also, it is infeasible to manually obtain segment annotations for retail stores like Amazon with millions of different products. 
Unfortunately, fully supervised neural networks cannot be applied without aspect labels. 
Moreover, the topics learned by unsupervised neural topic models are not perfectly aligned with the users' aspects of interest, so substantial human effort is required for interpreting and mapping the learned topics to meaningful aspects. 

Here, we investigate whether neural networks can be effectively trained under this challenging setting when only a small number of descriptive keywords, or \emph{seed words}, are available for each aspect class.
Table~\ref{tab:seed-word-examples} shows examples of aspects and five of their corresponding seed words from our experimental datasets (described later in more detail).
In contrast to a classification label, which is only relevant for a single segment, a seed word can implicitly provide aspect supervision to potentially many segments. 
We assume that the seed words have already been collected either manually or automatically.
Indeed, collecting a small\footnote{In our experiments, we only consider around 30 seed words per aspect. For comparison, the vocabulary of the datasets has more than 10,000 terms.} 
set of seed words per aspect is typically easier than manually annotating thousands of segments for training neural networks.
As we will see, even noisy seed words that are only weakly predictive of the aspect will be useful for aspect detection.

\begin{table}[t]
    \centering
    \resizebox{\columnwidth}{!}{
    \begin{tabular}{|l|l|}
    \hline
       \textbf{Aspect}  & \textbf{Seed Words} \\\hline
    Price (EN)     & price, value, money, worth, paid \\
    Image (EN)     & picture, color, quality, black, bright \\
    Food (EN) & food, delicious, pizza, cheese, sushi \\
    Drinks (FR) & vin, bière, verre, bouteille, cocktail\\
    Ambience (SP) & ambiente, mesas, terraza, acogedor, ruido\\
    \hline
    \end{tabular}}
    \caption{Examples of aspects and five of their corresponding seed words in various domains (electronic products, restaurants) and languages (``EN'' for English, ``FR'' for French, ``SP'' for Spanish).}
    \label{tab:seed-word-examples}
\end{table}
Training neural networks for segment-level aspect detection using just a few seed words is a challenging task. 
Indeed, as a contribution of this paper, we observe that current weakly supervised networks do not effectively leverage the predictive power of the available seed words. 
To address the shortcomings of previous seed word-based approaches, we propose a novel \emph{weakly supervised} approach, which uses the available seed words in a more effective way. 
In particular, we consider a \emph{student-teacher} framework, according to which a bag-of-seed-words classifier (teacher) is applied on unlabeled segments to supervise a second model (student), which can be any supervised model, including neural networks.

Our approach introduces several important contributions. 
First, our teacher model considers each individual seed word as a (noisy) aspect indicator, which as we will show, is more effective than previously proposed weakly supervised approaches. 
Second, by using only the teacher's aspect probabilities, our student generalizes better than the teacher and, as a result, the student outperforms both the teacher and previously proposed weakly supervised models. 
Finally, we show how iterative co-training can be used to cope with noisy seed words: the teacher effectively estimates the predictive quality of the noisy seed words in an unsupervised manner using the associated predictions by the student. 
Iterative co-training then leads to both improved teacher and student models. 
Overall, our approach consistently outperforms existing weakly supervised approaches, as we show with an experimental evaluation over six domains of product reviews and six multilingual datasets of restaurant reviews.

The rest of this paper is organized as follows. 
In Section~\ref{s:background} we review relevant work. 
In Section~\ref{s:our-distillation-approach} we describe our proposed weakly supervised approach.
In Section~\ref{s:aspect-extraction-experiments} we present our experimental setup and findings.
Finally, in Section~\ref{s:conclusion} we conclude and suggest future work. 
A preliminary version of this work was presented at the Second Learning from Limited Labeled Data Workshop~\cite{karamanolakis2019seedwords}.

\section{Related Work and Problem Definition}
\label{s:background}
We now review relevant work on aspect detection (Section~\ref{s:related-work-aspect-detection}), co-training (Section~\ref{s:co-training}), and knowledge distillation (Section~\ref{s:background-knowledge-distillation}). 
We also define our problem of focus (Section~\ref{s:problem-definition}).

\subsection{Segment-Level Aspect Detection}
\label{s:related-work-aspect-detection}
The goal of segment-level aspect detection is to classify a segment $s$ to $K$ aspects of interest.

\paragraph{Supervised Approaches.}
\label{s:supervised-aspect-extraction}
Rule-based or traditional learning models for aspect detection have been outperformed by supervised neural networks~\citep{liu2015fine,poria2016aspect,zhang2018deep}.
Supervised neural networks first use an embedding function\footnote{Examples of segment embedding functions are the average of word embeddings~\cite{wieting2015towards,arora2016simple}, Recurrent Neural Networks (RNNs)~\cite{yang2016hierarchical,wieting2017revisiting}, Convolutional Neural Networks (CNNs)~\cite{kim2014convolutional}, self-attention blocks~\cite{devlin2019bert,radford2018improving}, etc.} ($\operatorname{EMB}$) to compute a low dimensional segment representation $h = \operatorname{EMB}(s) \in \mathbb{R}^d$ and then feed $h$ to a classification layer\footnote{The classification layer is usually a hidden layer followed by the softmax function.} ($\operatorname{CLF}$) to predict probabilities for the $K$ aspect classes of interest: $p = \langle p^1, \dotsc, p^K \rangle = \operatorname{CLF}(h)$.
For simplicity, we write $p=f(s)$.
The parameters of the embedding function and the classification layer are learned using ground truth, segment-level aspect labels. 
However, aspect labels are not available in our setting, which hinders the application of supervised learning approaches. 

\paragraph{Unsupervised Approaches.}
Topic models have been used to train aspect detection with unannotated documents.
Recently, neural topic models~\citep{iyyer2016feuding,srivastava2017autoencoding,he2017unsupervised} have been shown to produce more coherent topics than earlier models such as Latent Dirichlet Allocation (LDA)~\citep{blei2003latent}.
In their Aspect Based Autoencoder (ABAE),~\citet{he2017unsupervised} first use segment $s$ to predict aspect probabilities $p=f(s)$ and then use $p$ to reconstruct an embedding $h'$ for $s$ as a convex combination of $K$ aspect embeddings: $h' = \sum_{k=1}^K p^k A_k$, where $A_k \in \mathbb{R}^d$ is the embedding of the $k$-th aspect.
The aspect embeddings $A_k$ are initialized by clustering the vocabulary embeddings using k-means with $K$ clusters.
ABAE is trained by minimizing the segment reconstruction error.\footnote{The reconstruction error can be efficiently estimated using contrastive max-margin objectives~\citep{weston2011wsabie,pennington2014glove}.}

Unfortunately, unsupervised topic models are not effective when used directly for aspect detection.
In particular, in ABAE, the $K$ topics learned to reconstruct the segments are not necessarily aligned with the $K$ aspects of interest.
A possible fix is to first learn $K' >> K$ topics and do a $K'$-to-$K$ mapping as a post-hoc step.
However, this mapping requires either aspect labels or substantial human effort for interpreting topics and associating them with aspects. 
This mapping is nevertheless not possible if the learned topics are not aligned with the aspects.

\paragraph{Weakly Supervised Approaches.}
Weakly supervised approaches use minimal domain knowledge (instead of ground truth labels) to model meaningful aspects.
In our setting, domain knowledge is given as a set of seed words for each aspect of interest~\cite{lu2011multi,lund2017tandem,angelidis2018summarizing}.
\citet{lu2011multi} use seed words as asymmetric priors in probabilistic topic models (including LDA).
\citet{lund2017tandem} use LDA with fixed topic-word distributions, which are learned using seed words as ``anchors'' for topic inference~\cite{arora2013practical}.
Neither of these two approaches can be directly applied into more recent neural networks for aspect detection. 
\citet{angelidis2018summarizing} recently proposed a weakly supervised extension of the unsupervised ABAE. 
Their model, named Multi-seed Aspect Extractor, or MATE, initializes the aspect embedding $A_k$ using the weighted average of the corresponding seed word embeddings (instead of the k-means centroids). %
To guarantee that the aspect embeddings will still be aligned with the $K$ aspects of interest after training,~\citet{angelidis2018summarizing} keep the aspect and word embeddings fixed throughout training.
In this work, we will show that the predictive power of seed words can be leveraged more effectively by considering each individual seed word as a more direct source of supervision during training. 

\subsection{Co-training}
\label{s:co-training}
Co-training~\cite{blum1998combining} is a classic multi-view learning method for semi-supervised learning.
In co-training, classifiers over different feature spaces are encouraged to agree in their predictions on a large pool of unlabeled examples. 
\citet{blum1998combining} justify co-training in a setting where the different views are conditionally independent given the label. Several subsequent works have relaxed this assumption and shown co-training to be effective in much more general settings~\cite{balcan2005co,chen2011co,collins1999unsupervised,clark2018semi}.
Co-training is also related to self-training (or bootstrapping)~\cite{yarowsky1995unsupervised}, which trains a classifier using its own predictions and has been successfully applied for various NLP tasks~\cite{collins1999unsupervised,mcclosky2006effective}.

Recent research has successfully revisited these general ideas to solve NLP problems with modern deep learning methods.
\citet{clark2018semi} propose ``cross-view training'' for sequence modeling tasks by modifying Bi-LSTMs for \emph{semi-supervised} learning.
\citet{ruder2018strong} show that classic bootstrapping approaches such as tri-training~\cite{zhou2005tri} can be effectively integrated in neural networks for semi-supervised learning under domain shift.
Our work provides further evidence that co-training can be effectively integrated into neural networks and combined with recent transfer learning approaches for NLP~\cite{dai2015semi,howard2018universal,devlin2019bert,radford2018improving}, in a substantially different, \emph{weakly supervised} setting where no ground-truth labels but only a few seed words are available for training.

\subsection{Knowledge Distillation}
\label{s:background-knowledge-distillation}
Our approach is also related to the ``knowledge distillation'' framework~\cite{bucilua2006model,ba2014deep,hinton2015distilling}, which has received considerable attention recently~\cite{lopez2015unifying, kim2016sequence, furlanello2018born,wang2019everything}.
Traditional knowledge distillation aims at compressing a cumbersome model (teacher) to a simpler model (student) by training the student using both ground truth labels and the soft predictions of the teacher in a distillation objective. 
Our work also considers a student-teacher architecture and the distillation objective but under a considerably different, weakly supervised setting: 
(1) we do not use any labels for training and (2) we create conditions that allow the student to outperform the teacher; in turn, (3) we can use the student’s predictions to learn a better teacher under co-training.

\begin{table}[]
    \centering
    \resizebox{\columnwidth}{!}{
    \begin{tabular}{|l|l|}
    \hline 
        Variable & Description  \\\hline
   $s$         & Segment (e.g., sentence) of a text review  \\
   $K$         & Number of aspects of interest \\
   $D$         & Total number of seed words \\
   $G_i \quad (i=1, \ldots, K)$ & Set of seed words for the $i$-th aspect\\
   $h \in \mathbb{R}^d$ & Segment embedding (student)\\
   $c \in \mathbb{N}^D$ & Bag-of-seed-words representation of $s$\\\hline
   $p=\langle p^1,\dots,p^K\rangle$ & Student's aspect predictions\\
   $q=\langle q^1,\dots,q^K\rangle$ & Teacher's aspect predictions\\
  \hline
    \end{tabular}}
    \caption{Notation.}
    \label{tab:notation}
\end{table}
\subsection{Problem Definition}
\label{s:problem-definition}
Consider a corpus of text reviews from an entity domain (e.g., televisions, restaurants).
Each review is split into segments (e.g., sentences, clauses). %
We also consider $K$ pre-defined aspects of interest $(1,\dots,K)$, including the ``General'' aspect, which we assume is the $K$-th aspect for simplicity.
Different segments of the same review may be associated with different aspects but ground-truth aspect labels are \emph{not} available for training. 
Instead, a small number of seed words $G_k$ are provided for each aspect $k \in [K]$.
Our goal is to use the corpus of training reviews and the available seed words $G=(G_1, \dots, G_K)$ to train a classifier, which, given an unseen test segment $s$, predicts $K$ aspect probabilities $p=\langle p^1,\dots,p^K\rangle$. 

\section{Our Student-Teacher Approach}
\label{s:our-distillation-approach}
\begin{figure}[t]
\centering
\includegraphics[scale=0.38]{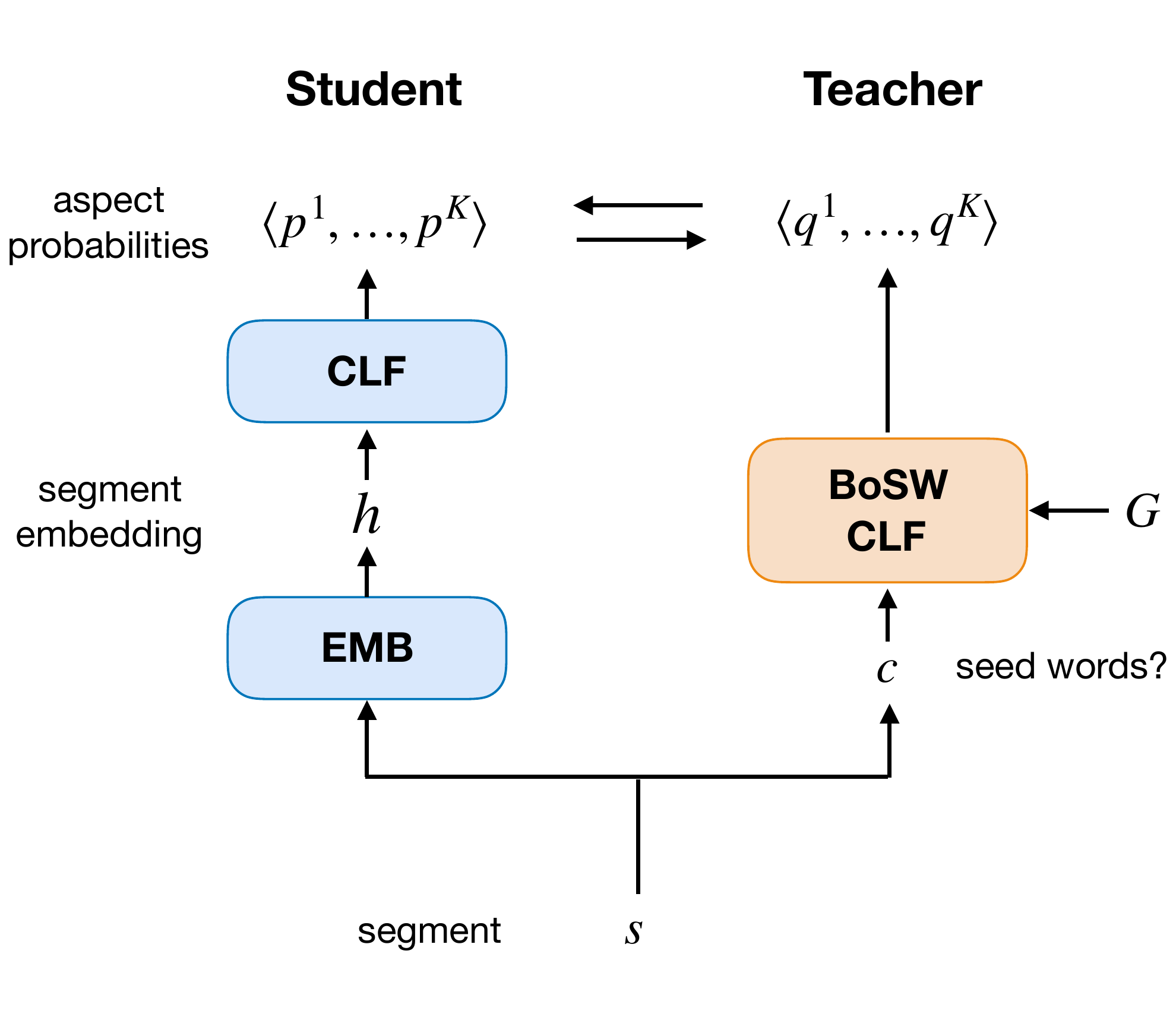}
	\caption{Our student-teacher approach for segment-level aspect detection using seed words. }
\label{fig:student_teacher_architecture}
\end{figure}

We now describe our weakly supervised framework for aspect detection.
We consider a student-teacher architecture (Figure~\ref{fig:student_teacher_architecture}), where the teacher is a bag-of-words classifier based solely on the provided seed words (i.e., a ``bag-of-seed-words'' classifier), and the student is an embedding-based neural network trained on data ``softly'' labeled by the teacher (as in the distillation objective).
In the rest of this section, we describe the individual components of our student-teacher architecture and our proposed algorithm for performing updates.

\subsection{Teacher: A Bag-of-Seed-Words Classifier}
\label{s:teacher}
Our teacher model leverages the available seed words $G$ that are predictive of the $K$ aspects.
Let $D$ denote the total number of seed words in $G$.
We can represent a segment $s_i$ using a bag-of-seed-words representation $c_i \in \mathbb{N}^D$, where $c_i^j$ encodes the number of times the $j$-th seed word occurs in $s_i$. (Note that $c_i$ ignores the non-seed words.)
The teacher's prediction for the $k$-th aspect is: 
\begin{equation}
    q_i^k = \frac{\exp(\sum_{j=1}^D \mathbbm{1}\{j \in G_k\} \cdot c_i^j)}{\sum_{k'} \exp (\sum_{j=1}^D \mathbbm{1}\{j \in G_{k'}\} \cdot c_i^j)}.
        \label{eq:teacher-probability-without-quality}
\end{equation}
If no seed word appears in $s$, then the teacher predicts the ``General'' aspect by setting $q_i^K=1$. 
Under this configuration the teacher uses seed words in a direct and intuitive way: it predicts aspect probabilities for the $k$-th aspect, which are proportional to the counts of the seed words under $G_k$, while if no seed word occurs in $s$, it predicts the ``General'' aspect.
The classifier receives $c_i$ as input and predicts $q_i=\langle q_i^1,\dots,q_i^K \rangle$. 

Although the teacher only uses seed words to predict the aspect of a segment, we also expect non-seed words to carry predictive power.
Next, we describe the student network that learns to associate non-seed words with aspects. 

\subsection{Student: An Embedding-Based Network}
\label{s:student-network}
Our student model is an embedding-based neural network: a segment is first embedded ($h_i = \operatorname{EMB}(s_i) \in \mathbb{R}^d$) and then classified to the $K$ aspects ($p_i = \operatorname{CLF}(h_i)$) (see Section~\ref{s:supervised-aspect-extraction}). 
The student does not use ground-truth aspect labels for training. 
Instead, it is trained by optimizing the distillation objective, i.e., the cross entropy between the teacher's (soft) predictions and the student's predictions: 
\begin{equation}
    H(q_i, p_i) = - \sum_k q_i^k \log p_i^k
    \label{eq:distillation-loss}
\end{equation}
While the teacher only uses the seed words in $s_i$ to form its prediction $q_i$, the student uses all the words in $s_i$. 
Thus, using the distillation loss for training, the student learns to use both seed words and non-seed words to predict aspects.
As a result, the student is able to generalize better than the teacher and \emph{predict aspects even in segments that do not contain any seed words}. 
To regularize the student model, we apply L2 regularization to the classifier's weights and dropout regularization to the word embeddings~\cite{srivastava2014dropout}. 
As we will show in Section~\ref{s:aspect-extraction-experiments}, our student with this configuration outperforms the teacher in aspect prediction. 

\subsection{Iterative Co-Training}
\label{eq:iterative-co-training}
In this section, we describe our iterative co-training algorithm to cope with noisy seed words. 
The teacher in Section~\ref{s:teacher} considers each seed word equally, which can be problematic because not all seed words are equally good for predicting an aspect.
In this work, we propose to estimate the predictive quality of each seed word in an unsupervised way. 
Our approach is inspired in the Model Bootstrapped Expectation Maximization (MBEM) algorithm of~\citet{khetan2017learning}.
MBEM is guaranteed to converge (under mild conditions) when the number of training data is sufficiently large and the worker quality is sufficiently high.
Here, we treat seed words as ``noisy annotators'' and adopt an iterative estimation procedure similar to MBEM, as we describe next.

We model the predictive quality of the $j$-th seed word as a weight vector $z_j=\langle z_j^1, \dots, z_j^K \rangle$, where $z_j^k$ measures the strength of the association with the $k$-th aspect.
We thus change the teacher to consider seed word quality.
In particular, we replace Equation~\eqref{eq:teacher-probability-without-quality} by: 
\begin{equation}
    q_i^k = \frac{\exp \sum_{j=1}^D \mathbbm{1}\{j \in G_k\} \cdot  \hat z_j^k \cdot c_i^j}{\sum_{k'} \exp  \sum_{j=1}^D  \mathbbm{1}\{j \in G_{k'}\} \cdot \hat z_j^{k'} \cdot c_i^j},
    \label{eq:teacher-probability-with-quality}
\end{equation}
where $\hat z_j$ is the current estimate of $z_j$. 
As no ground-truth labels are available, we follow~\citet{khetan2017learning} and estimate $z_j$ via Maximum Likelihood Estimation using the student's predictions as the current estimate of the ground truth labels. 
In particular, we assume that the prediction of the student for a training segment $s_i$ is $t_i = \operatorname{argmax}_k p_i^k $. Then, for each seed word we compute the quality estimate for the $k$-th aspect using the student's predictions for $N$ segments:
\begin{equation}
    \hat z_j^k = \frac{\sum_{i=1}^N \mathbbm{1}\{c _i^j > 0\} \mathbbm{1}\{t_i = k\} }{\sum_{k'} \sum_{i=1}^N  \mathbbm{1}\{c _i^j > 0\} \mathbbm{1}\{t_i = k'\}} .
    \label{eq:teacher-update}
\end{equation}
 According to Equation~\eqref{eq:teacher-update}, the quality of the $j$-th seed word is estimated according to the student-teacher agreement on segments where the seed word appears.

\begin{figure}
\centering
\begin{algorithm}[H]
  \begin{algorithmic}
    \State{\textbf{Input:} $\{s_i\}_{i\in [N]}$, $D$ seed words grouped into $K$ disjoint sets $G=(G_1,\dots, G_K)$}
    \State{\textbf{Output:} $\hat f$: predictor function for segment-level aspect detection\\\hrulefill}
  \end{algorithmic}
  \begin{algorithmic}
    \State{Predict $\{q_i\}_{i\in[N]}$ (Eq.~\eqref{eq:teacher-probability-without-quality})}
    \Comment{\textit{Apply teacher}}
    \State{\textbf{Repeat until convergence criterion}}
      \begin{algorithmic}
      \State{Learn $\hat f$ (Eq.~\eqref{eq:distillation-loss})}
      \Comment{\textit{Train student}}
      \State{Predict $\{p_i=\hat f(s_i)\}_{i\in[N]}$}
      \Comment{\textit{Apply student}}
      \State{Update $\{z_j\}_{j \in [D]}$ (Eq.~\eqref{eq:teacher-update})}
      \Comment{\textit{Update teacher}}
      \State{Predict $\{q_i\}_{i\in[N]}$ (Eq.~\eqref{eq:teacher-probability-with-quality})}
      \Comment{\textit{Apply teacher}}
    \end{algorithmic}
  \end{algorithmic}
  \caption*{\textbf{Algorithm 1} Iterative Seed Word Distillation}
  \label{algo:iswd}
\end{algorithm}
\end{figure}

Building upon the previous ideas, we present our Iterative Seed Word Distillation (ISWD) algorithm for effectively leveraging the seed words for fine-grained aspect detection. 
Each round of ISWD consists of the following steps (Algorithm 1): 
(1) we apply the teacher on unlabeled training segments to get predictions $q_i$ (without considering seed word qualities);
(2) we train the student using the teacher's predictions in the distillation objective of Equation~\eqref{eq:distillation-loss};\footnote{Note that the quality-aware loss function proposed in~\citet{khetan2017learning}, which is an alternative form of noise-aware loss functions~\cite{natarajan2013learning}, is equivalent to our distillation loss: using the log loss as $l(.)$ in Equation (4) of~\citet{khetan2017learning} yields the cross entropy loss.}
(3) we apply the student in the training data to get predictions $p_i$; and
(4) we update the seed word quality parameters using the student's predictions in Equation~\eqref{eq:teacher-update}.

In contrast to MATE, which uses the validation set (with aspect labels) to estimate seed weights in an initialization step, our proposed method is an unsupervised approach to modeling and adapting the seed word quality during training.
We stop this iterative procedure after the disagreement between the student's and teacher's hard predictions in the training data stops decreasing. 
We empirically observe that 2-3 rounds are sufficient to satisfy this criterion.
This observation also agrees with~\citet{khetan2017learning}, who only run their algorithm for two rounds.

\section{Experiments}
\label{s:aspect-extraction-experiments}
We evaluate our approach to aspect detection on several datasets of product and restaurant reviews.

\subsection{Experimental Settings}
\paragraph{Datasets.} 
We train and evaluate our models on Amazon product reviews for six domains (Laptop Bags, Keyboards, Boots, Bluetooth Headsets, Televisions, and Vacuums) from the OPOSUM dataset~\cite{angelidis2018summarizing}, and on restaurant reviews in six languages (English, Spanish, French, Russian, Dutch, Turkish) from the SemEval-2016 Aspect-based Sentiment
Analysis task~\cite{pontiki2016semeval}. 
Aspect labels (9-class for product reviews and 12-class for restaurant reviews) are available for each segment\footnote{In product reviews, elementary discourse units (EDUs) are used as segments. In restaurant reviews, sentences are used as segments.} of the validation and test sets.
The restaurant reviews also come with training aspect labels, which we only use for training the fully supervised models.
For a fair comparison, we use exactly the same 30 seed words (per aspect and domain) used in~\citet{angelidis2018summarizing} for the product reviews and use the same extraction method described in~\citet{angelidis2018summarizing} to extract 30 seed words for the restaurant reviews.
See Appendix~\ref{s:appendix} for more dataset details.

\paragraph{Experimental Procedure.} 
For a fair comparison, we use exactly the same pre-processing (tokenization, stemming, and word embedding) and evaluation procedure as in~\citet{angelidis2018summarizing}. 
For each domain, we train our model on the training set without using any aspect labels, and only use the seed words $G$ via the teacher. 
For each model, we report the average test performance over 5 different runs with the parameter configuration that achieves best validation performance. 
As evaluation metric, we use the micro-averaged F1.

\paragraph{Model Configuration.} 
For the student network, we experiment with various modeling choices for segment representations: bag-of-words (BOW) classifiers, the unweighted average of word2vec embeddings (W2V), the weighted average of word2vec embeddings using bilinear attention~\cite{luong2015effective} (same setting as~\citet{he2017unsupervised,angelidis2018summarizing}), and the average of contextualized word representations obtained from the second-to-last layer of the pre-trained (self-attention based) BERT model~\cite{devlin2019bert}, which uses multiple self-attention layers~\cite{vaswani2017attention} and has been shown to achieve state-of-the-art performance in many downstream NLP applications. 
For the English product reviews, we use the base uncased BERT model. For the multilingual restaurant reviews, we use the multilingual cased BERT model.\footnote{Both models can be found in \href{https://github.com/google-research/bert/blob/master/multilingual.md}{https://github.com/google-research/bert/blob/master/multilingual.md}. The multilingual cased BERT model is recommended by the authors instead of the multilingual uncased BERT model.}

In iterative co-training, we train the student network to convergence in each iteration (which may require more than one epoch over the training data).
Moreover, we observed that the iterative process is more stable when we interpolate between weights of the previous iteration and the estimated updates instead of directly applying the estimated seed weight updates (according to Equation~\eqref{eq:teacher-probability-with-quality}).

\begin{table*}[t]
\centering
\resizebox{2\columnwidth}{!}{
\begin{tabular}{|l||cccccc|c|}
\hline 
   &      \multicolumn{6}{c|}{\textbf{Product Review Domain}}  &  \multicolumn{1}{c|}{}      \\ 

 \textbf{Method} & \textbf{Bags} & \textbf{Keyboards} & \textbf{Boots} & \textbf{Headsets} & \textbf{TVs}  & \textbf{Vacuums} & \textbf{AVG}       \\
 \hline\hline
LDA-Anchors~\cite{lund2017tandem} & 33.5   & 34.7  & 31.7 & 38.4  & 29.8  & 30.1 & 33.0 \\
ABAE~\cite{he2017unsupervised}                             & 38.1   & 38.6  & 35.2    & 37.6 & 39.5  & 38.1 & 37.9 \\
\hline
MATE~\cite{angelidis2018summarizing}                             & 46.2   & 43.5  & 45.6    & 52.2 & 48.8  & 42.3 & 46.4 \\
MATE-unweighted & 41.6   & 41.3  & 41.2    & 48.5 & 45.7  & 40.6 & 43.2 \\
MATE-MT (best performing)                         & 48.6   & 45.3  & 46.4    & 54.5 & 51.8  & 47.7 & 49.1 \\
\hline
Teacher & 55.1   & 52.0  & 44.5    & 50.1   & 56.8  & 54.5 & 52.2 \\
\hline
Student-BoW                 & 57.3   & 56.2    & 48.8    & 59.8 & 59.6  & 55.8 & 56.3 \\
Student-W2V                 &59.3   & 57.0  & 48.3    & \textbf{66.8} & \textbf{64.0}  & 57.0 & 58.7 \\
Student-W2V-RSW             & 51.3   & 57.2  & 46.6    & 63.0 & 62.1  & 57.1 & 56.2 \\

Student-ATT  &60.1	&55.6	&49.9	&66.6	& 63.4 &	58.2	& 58.9 \\
Student-BERT             & \textbf{61.4}   & \textbf{57.5}  & \textbf{52.0}    & 66.5 & 63.0  & \textbf{60.4} & \textbf{60.2} \\
\hline
\end{tabular}}
\caption{Micro-averaged F1 reported for 9-class EDU-level aspect detection in product reviews.}
\label{tab:aspect-extraction-results-oposum}
\end{table*}

\begin{table*}[t]
\centering
\resizebox{1.35\columnwidth}{!}{
\begin{tabular}{|c||cccccc|c|}
\hline 
          &      \multicolumn{6}{c|}{\textbf{Restaurant Review Language}}  &  \multicolumn{1}{c|}{}      \\ 

\textbf{Method}  & \textbf{En}                          & \textbf{Sp} & \textbf{Fr} & \textbf{Ru} & \textbf{Du}  & \textbf{Tur} & \textbf{AVG}       \\
 \hline\hline
W2V-Gold & 58.8 & 50.4 & 50.4 & 69.3 & 51.4 & 55.7 & 56.0 \\
BERT-Gold & 63.1 & 51.6 & 50.5 & 64.6 & 53.5 & 55.3 & 56.4 \\
\hline
MATE &	41.0 &	24.9 &	17.8	& 18.4 &	36.1	 & 39.0 & 	29.5 \\
MATE-unweighted& 40.3	 & 18.3 &	19.2& 21.8  &	 31.5 & 25.2 & 26.1  \\

\hline 
Teacher	& 44.9 &	41.8	 & 34.1& 	54.4 &	40.7 &	30.2	& 41.0 \\
\hline 
Student-W2V	& 47.2 &	40.9	& 32.4 &	\textbf{59.0}	 & 42.1 &	42.3 &	44.0 \\
Student-ATT	& 47.8	& 41.7 &	32.9 &	57.3 &	44.1 &	45.5 &	44.9 \\
Student-BERT &	\textbf{51.8} &	\textbf{42.0}	& \textbf{39.2} &	58.0	& \textbf{43.0}	& \textbf{45.0}	& \textbf{46.5} \\
\hline
\end{tabular}}
\caption{Micro-averaged F1 reported for 12-class sentence-level aspect detection in restaurant reviews. The fully supervised *-Gold models are not directly comparable with the weakly supervised models.}
\label{tab:aspect-extraction-results-semeval}
\end{table*}

\paragraph{Model Comparison.} 
For a robust evaluation of our approach, we compare the following models and baselines:
\begin{itemize}
    \item \textbf{LDA-Anchors:} The topic model of~\citet{lund2017tandem} using seed words as ``anchors.''
    \item \textbf{ABAE:} The unsupervised autoencoder of~\citet{he2017unsupervised}, where the learned topics were manually mapped to aspects.
    \item \textbf{MATE-*:}  The MATE model of~\citet{angelidis2018summarizing} with various configurations: initialization of the aspect embeddings $A_k$ using the unweighted/weighted average of seed word embeddings and an extra multi-task training objective (MT).\footnote{The multi-task training objective in MATE requires datasets from different domains but same language, thus it cannot be applied in our datasets of restaurant reviews.} 
    \item \textbf{Teacher:} Our bag-of-seed-words teacher. 
    \item \textbf{Student-*:} Our student network trained with various configurations for the EMB function.
    \item \textbf{*-Gold:} Supervised models trained using ground truth aspect labels, which are only available for restaurant reviews. These models are not directly comparable with the other models and baselines. 
\end{itemize}

\subsection{Experimental Results}
Tables~\ref{tab:aspect-extraction-results-oposum} and~\ref{tab:aspect-extraction-results-semeval} show the results for aspect detection on product and restaurant reviews, respectively.
The rightmost column of each table reports the average performance across the 6 domains/languages.  

\paragraph{MATE-* models outperform ABAE.} Using the seed words to initialize aspect embeddings leads to more accurate aspect predictions than mapping the learned (unsupervised) topics to aspects.

\paragraph{LDA-Anchors performs worse than MATE-* models.} Although averages of seed words were used as ``anchors'' in the ``Tandem Anchoring'' algorithm, we observed that the learned topics did not correspond to our aspects of interest.

\paragraph{The teacher effectively leverages seed words.}
By leveraging the seed words in a more direct way, Teacher is able to outperform the MATE-* models.
Thus, we can use Teacher's predictions as supervision for the student, as we describe next.

\paragraph{The student outperforms the teacher.}
Student-BoW outperforms Teacher: the two models have the same architecture but Teacher only considers seed words; regularizing Student's weights encourages Student to mimic the noisy aspect predictions of Teacher by also considering non-seed words for aspect detection. 
The benefits of our distillation approach are highlighted using neural networks with word embeddings. 
Student-W2V outperforms both Teacher and Student-BoW, showing that obtaining segment representations as the average of word embeddings is more effective than using bag-of-words representations for this task. 

\paragraph{The student outperforms previous weakly supervised models even in one co-training round.}
Student-ATT outperforms MATE-unweighted (by 36.3\% in product reviews and by 52.2\% in restaurant reviews) even in a single co-training round: although the two models use exactly the same seed words (without weights), pre-trained word embeddings, EMB function, and CLF function, our student-teacher approach leverages the available seed words more effectively as noisy supervision than just for initialization. 
Also, using our approach, we can explore more powerful methods for segment embedding without the constraint of a fixed word embedding space.
Indeed, using contextualized word representations in Student-BERT leads to the best performance over all models.

As expected, our weakly supervised approach does not outperform the fully supervised (*-Gold) models. 
However, our approach substantially reduces the performance gap between weakly supervised approaches and fully supervised approaches by 62\%.
The benefits of our student-teacher approach are consistent across all datasets, highlighting the predictive power of seed words across different domains and languages. 

\begin{figure}[t]
\centering
\includegraphics[scale=0.35]{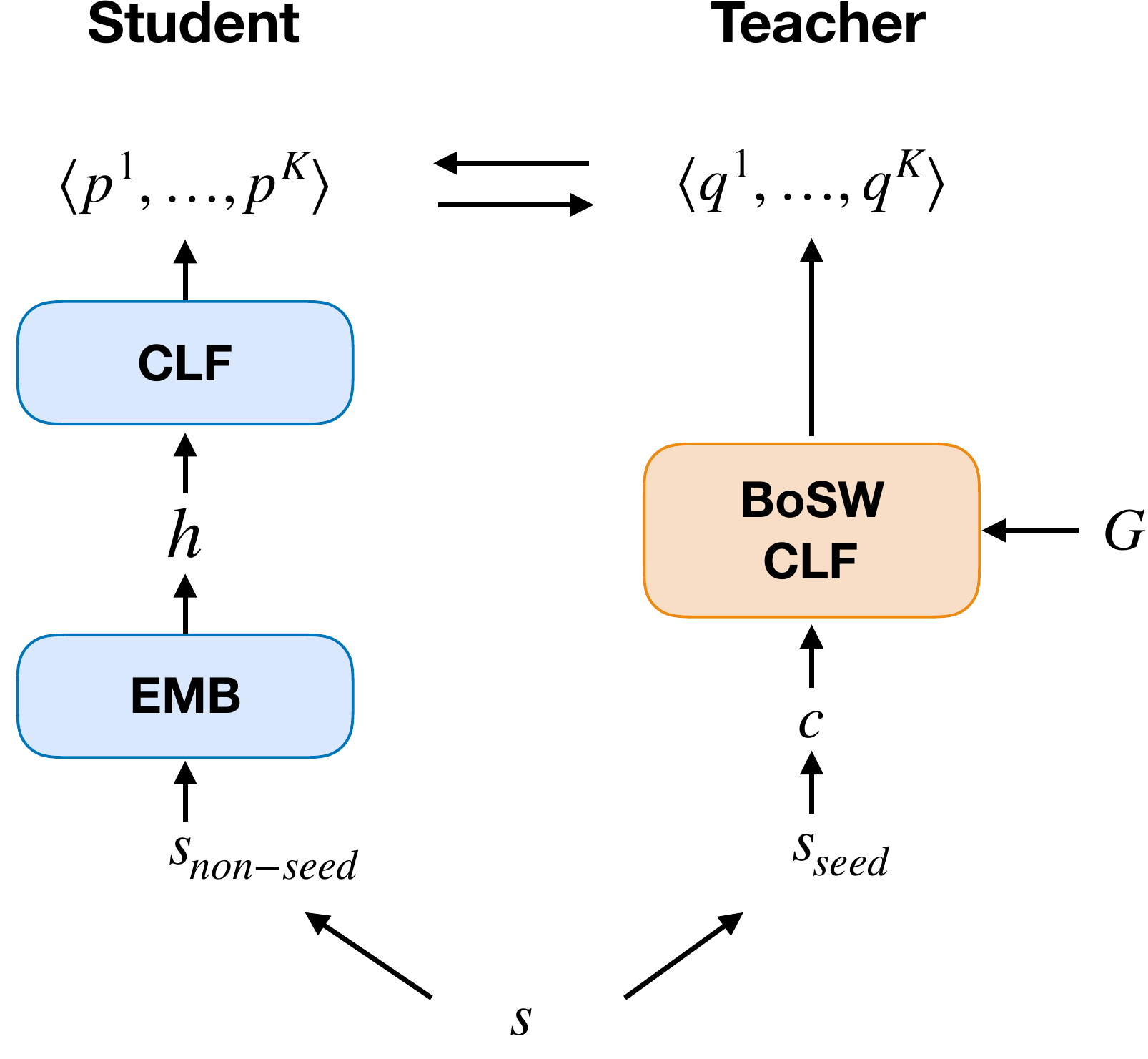}
	\caption{Our weakly supervised co-training approach when seed words are removed from the student's input (RSW baseline). Segment $s_{non-seed}$ is an edited version of $s$, where we replace each seed word in $s$ by an ``UNK'' special token (like out-of-vocabulary words).}
\label{fig:rsw_baseline}
\end{figure}

\paragraph{The student leverages non-seed words.}
To better understand the extent to which non-seed words can predict the aspects of interest, we experiment with completely removing the seed words from Student-W2V's input during training (Student-W2V-RSW method; see Figure~\ref{fig:rsw_baseline}). %
Thus, in this setting, Student-W2V-RSW is forced to only use non-seed words to detect aspects.
Note that the co-training assumption of conditionally independent views~\cite{blum1998combining} is satisfied in this setting, where Teacher is only using seed words and Student-W2V is only using non-seed words. 
Student-W2V-RSW effectively learns to use non-seed words to predict aspects and performs better than Teacher (but worse than Student-W2V, which considers both seed and non-seed words).
For additional ablation experiments, see Appendix~\ref{s:appendix}.

\begin{table}[t]
\centering
\resizebox{\columnwidth}{!}{
\begin{tabular}{|c|c|c|}
\hline 
 \textbf{Method}  & \textbf{Initial} & \textbf{Iterative}\\
 \hline\hline
 \multicolumn{3}{|c|}{Product Reviews (AVG)}  \\ 
\hline
MATE & 46.4 & - \\
Teacher / Student-W2V &  52.2 / 58.7 & \textbf{58.5} / \textbf{59.7} \\
Teacher / Student-BERT & 52.2 / 60.2 & \textbf{58.6} / \textbf{60.8}\\
\hline 
 \multicolumn{3}{|c|}{Restaurant Reviews (En)}  \\ 
\hline
MATE & 29.5 & - \\
Teacher / Student-W2V &  44.9 / 47.2 &	\textbf{45.8} / \textbf{49.0} \\
Teacher / Student-BERT &  44.9 / 51.8 &	\textbf{49.8} / \textbf{53.4} \\
\hline
\end{tabular}}
\caption{Micro-averaged F1 scores during the first round (middle column) and after iterative co-training (right column) in product reviews (top) and restaurant reviews (bottom). 
}
\label{tab:cotraining-results}
\end{table}

\paragraph{Iterative co-training copes with noisy words.}
Further performance improvement in Teacher and Student-* can be observed with the iterative co-training procedure of Section~\ref{eq:iterative-co-training}. 
Table~\ref{tab:cotraining-results} reports the performance of Teacher and Student-* after co-training for both product reviews (top) and English restaurant reviews (bottom). 
(For more detailed, per-domain results, see Appendix~\ref{s:appendix}.)
Compared to the initial version of Teacher that does not model the quality of the seed words, iterative co-training leads to estimates of seed word quality that improve Teacher's performance up to 12.3\% (in product reviews using Student-BERT).

\paragraph{A better teacher leads to a better student.}
Co-training leads to improved student performance in both datasets (Table~\ref{tab:cotraining-results}). 
Compared to MATE, which uses the validation set to estimate the seed weights as a pre-processing step, we estimate and iteratively adapt the seed weights using the student-teacher disagreement, which substantially improves performance.
Across the 12 datasets, Student-BERT leads to an average absolute increase of 14.1 F1 points. 

\begin{figure}[t]
  \begin{subfigure}[t]{0.5\columnwidth}
  \centering
    \includegraphics[height = 4.3cm, width = 4.1cm]{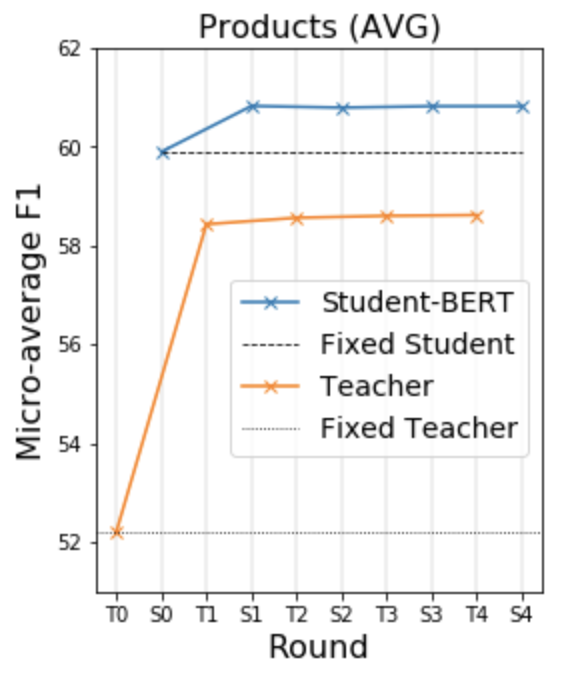}
  \end{subfigure}
    \begin{subfigure}[t]{0.4\columnwidth}
    \includegraphics[height = 4.52cm, width = 4cm]{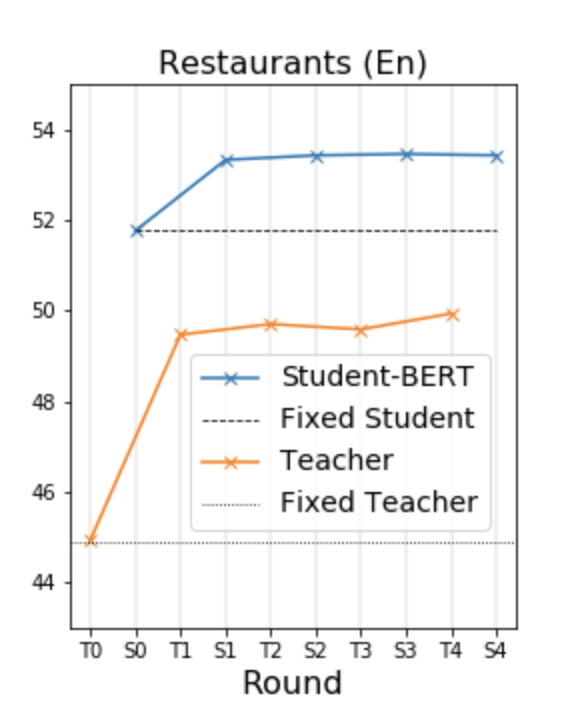}

  \end{subfigure}
  \caption{Co-training performance for each round reported for product reviews (left) and restaurant reviews (right). T\textless$i$\textgreater and S\textless$i$\textgreater  correspond to the teacher's and student's performance, respectively, at the $i$-th round.}
\label{fig:co-training-performance}
\end{figure}

Figure~\ref{fig:co-training-performance} plots Teacher's and Student-BERT's performance after each round of co-training. 
Most of the improvement for both Teacher and Student-BERT is gained in the first two rounds of co-training: ``T0'' (in Figure~\ref{fig:co-training-performance}) is the initial teacher, while ``T1'' is the teacher with estimates of seed word qualities, which leads to more accurate predictions, e.g., in segments with multiple seed words from different aspects.

\section{Conclusions and Future Work}
\label{s:conclusion}
We presented a weakly supervised approach for leveraging a small number of seed words (instead of ground truth aspect labels) for segment-level aspect detection. 
Our student-teacher approach leverages seed words more directly and effectively than previous weakly supervised approaches. 
The teacher model provides weak supervision to a student model, which generalizes better than the teacher by also considering non-seed words and by using pre-trained word embeddings.
We further show that iterative co-training lets us estimate the quality of the (possibly noisy) seed words. 
This leads to a better teacher and, in turn, a better student.
Our proposed method consistently outperforms previous weakly supervised methods in 12 datasets, allowing for seed words from various domains and languages to be leveraged for aspect detection. 
Our student-teacher approach could be applied for any classification task for which a small set of seed words describe each class.
In future work, we plan to extend our framework to multi-task settings, and to incorporate interaction to learn better seed words.

\subsubsection*{Acknowledgments}
We thank the anonymous reviewers for their constructive feedback. This material is based upon work supported by the National Science Foundation under Grant No. IIS-15-63785.

\bibliography{myreferences}
\bibliographystyle{acl_natbib}

\newpage 

\appendix
\section{Appendix}
\label{s:appendix}
For reproducibility, we provide more information on datasets (Section~\ref{s:appendix-datasets}) and implementation details (Section~\ref{eq:appendix-implementation-details}), and report more detailed evaluation results (Section~\ref{eq:appendix-more-results}).

\subsection{Datasets}
In this section, we describe all details of the datasets of product and restaurant reviews, and report dataset statistics. 
\label{s:appendix-datasets}
\paragraph{Product Reviews.} The OPOSUM dataset~\cite{angelidis2018summarizing} is a subset of the Amazon Product Dataset~\citep{mcauley2015image}, which contains Amazon reviews from 6 domains: Laptop Bags, Keyboards, Boots, Bluetooth Headsets, Televisions, and Vacuums. 
The validation and test segments of each domain have been manually annotated with 9 aspects (Table~\ref{tab:restaurant-review-aspects}). 
The reviews of each domain are already segmented by~\citet{angelidis2018summarizing} into elementary discourse units (EDUs) using a Rhetorical Structure Theory parser~\cite{feng2012text}.
The average number of training, validation, and test segments across domains is around $1$ million, $700$, and $700$ segments, respectively.  
Segment statistics per domain are reported in the supplementary material of~\citet{angelidis2018summarizing}.

\paragraph{Restaurant Reviews.} The datasets used in the SemEval-2016 Aspect-based Sentiment Analysis task~\cite{pontiki2016semeval} contain reviews for multiple domains and languages. 
Here, we use the six corpora of multilingual (English, Spanish, French, Russian, Dutch, Turkish) restaurant reviews.
The training, validation, and test segments have been manually annotated with 12 aspects, which are shared across languages: 
\begin{enumerate}
    \item  Restaurant\#General
    \item  Food\#Quality
    \item  Service\#General
    \item  Ambience\#General
    \item  Food\#Style\_Options
    \item  Food\#Prices
    \item  Restaurant\#Miscellaneous
    \item  Restaurant\#Prices
    \item  Drinks\#Quality
    \item  Drinks\#Style\_Options
    \item Location\#General
    \item Drinks\#Prices
\end{enumerate}
The reviews of each language are already segmented into sentences. 
The average number of training and test segments across languages is around 2500 and 800 segments respectively. 
The training segments of restaurant reviews are significantly fewer than the training segments of product reviews. 
Therefore, for non-English reviews we report results after a single co-training round. 
For our co-training experiments we augment the English reviews dataset with 50,000 English reviews randomly sampled from the Yelp Challenge corpus.\footnote{https://www.yelp.com/dataset/challenge}
\subsection{Implementation Details}
\label{eq:appendix-implementation-details}
For a fair comparison, for the product reviews we use the 200-dimensional word2vec embeddings provided by~\citet{angelidis2018summarizing} and the base uncased BERT model.\footnote{https://github.com/google-research/bert\#pre-trained-models}
For the restaurant reviews, we use the 300-dimensional multilingual word2vec embeddings provided by~\citet{ruder2016hierarchical} and the multilingual cased BERT model.\footnote{https://github.com/google-research/bert/blob/master/multilingual.md}
The student's parameters are optimized using Adam~\cite{kingma2014adam} with learning rate 0.005 and mini-batch size 50. 
After each co-training round we divide the learning rate by 10. 
We apply dropout in the word embeddings and the last hidden layers of the classifiers~\cite{srivastava2014dropout} with rate 0.5.

\subsection{More Results}
\label{eq:appendix-more-results}
Table~\ref{tab:aspect-extraction-results-oposum-detailed} reports detailed per-domain results. 
``Teacher (symmetric)'' is a simpler version of Teacher that randomly guesses the aspect of segments with no seed words. 
For Student-W2V we report additional ablation experiments. 
The *-ISWD models correspond to student or teacher models after multiple rounds of co-training until convergence. 

\begin{table*}[!htb]
    \renewcommand\thetable{4}

    \centering
    \resizebox{2\columnwidth}{!}{
    \begin{tabular}{|cccccc|}
    \hline
    \textbf{Bags} & \textbf{Keyboards} &  \textbf{Boots} &\textbf{Headsets}& \textbf{TVs} & \textbf{Vacuums}\\
    \hline
Size/Fit &Feel/Comfort& Comfort &Sound & Image & Accessories\\
Quality & Layout& Size&Comfort& Sound& Ease of Use\\
Looks & Build Quality& Look& Ease of Use & Connectivity & Suction Power\\
Compartments &Extra Function.& Materials& Connectivity & Customer Serv.& Build Quality\\
Handles & Connectivity& Durability& Durability & Ease of Use & Noise \\
Protection & Price& Weather Resist. &Battery & Price& Weight\\
Price & Noise& Price & Price& Apps/Interface & Customer Serv.\\
Customer Serv. &Looks& Color& Look & Size/Look& Price\\
General & General& General&General&General & General\\

    \hline
    \end{tabular}}
    \caption{The 9 aspect classes per domain of product reviews (OPOSUM).}
    \label{tab:restaurant-review-aspects}
\end{table*}

\begin{table*}[!htb]
\renewcommand\thetable{5}
\centering
\resizebox{2\columnwidth}{!}{
\begin{tabular}{|l||cccccc|c|}
\hline 
   &      \multicolumn{6}{c|}{\textbf{Product Review Domain}}  &  \multicolumn{1}{c|}{}      \\ 

 \textbf{Method} & \textbf{Bags} & \textbf{Keyboards} & \textbf{Boots} & \textbf{Headsets} & \textbf{TVs}  & \textbf{Vacuums} & \textbf{AVG}       \\
 \hline\hline
\multicolumn{8}{|c|}{Previous Approaches}\\
\hline
LDA-Anchors~\cite{lund2017tandem} & 33.5   & 34.7  & 31.7 & 38.4  & 29.8  & 30.1 & 33.0 \\
ABAE~\cite{he2017unsupervised}                             & 38.1   & 38.6  & 35.2    & 37.6 & 39.5  & 38.1 & 37.9 \\
\hline
MATE~\cite{angelidis2018summarizing}                             & 46.2   & 43.5  & 45.6    & 52.2 & 48.8  & 42.3 & 46.4 \\
MATE-unweighted & 41.6   & 41.3  & 41.2    & 48.5 & 45.7  & 40.6 & 43.2 \\
MATE-MT (best performing)                         & 48.6   & 45.3  & 46.4    & 54.5 & 51.8  & 47.7 & 49.1 \\
\hline
\multicolumn{8}{|c|}{Our Approach: Single Round Co-training}\\
\hline
Teacher (symmetric)	& 38.9 &	27.7 &	30.3 &	34.0 &	33.5 &	35.6 &	33.3\\
Teacher & 55.1   & 52.0  & 44.5    & 50.1   & 56.8  & 54.5 & 52.2 \\
\hline
Student-BoW                 & 57.3   & 56.2    & 48.8    & 59.8 & 59.6  & 55.8 & 56.3 \\
\hline
Student-W2V                 &59.3   & 57.0  & 48.3    & \textbf{66.8} & \textbf{64.0}  & 57.0 & 58.7 \\
Student-W2V-RSW             & 51.3   & 57.2  & 46.6    & 63.0 & 62.1  & 57.1 & 56.2 \\
Student-W2V w/o L2 Reg             & 56.3	& 56.6 &	48.8 &	59.8	& 58.4	& 54.7	& 55.7  \\
Student-W2V w/o dropout             & 56.4 &	56.2 &	48.1 &	59.4 & 57.4 &	54.2 &	55.3  \\
Student-W2V w/o emb fine-tuning            & 58.7 &	53.6	& 42.8 &	62.2 &	56.3 &	54.3 &54.6 \\
Student-W2V w/o soft targets             &57.2 &	57.4 &	47.1 &	61.7	&58.3 &	55.0	&56.1\\
\hline
Student-ATT  &60.1	&55.6	&49.9	&66.6	& 63.4 &	58.2	& 58.9 \\
Student-BERT             & \textbf{61.4}   & \textbf{57.5}  & \textbf{52.0}    & 66.5 & 63.0  & \textbf{60.4} & \textbf{60.2} \\

\hline
\multicolumn{8}{|c|}{Our Approach: Iterative Co-training}\\
\hline
Teacher-ISWD (St: W2V) &	59.3 &	58.2 &	50.6 &	63.6 &	61.0 &	58.4 &	58.5 \\
Teacher-ISWD (St: ATT) &	59.6 &	58.0 &	50.6 &	62.4 &	60.6 &	59.0	& 58.3\\
Teacher-ISWD (St: BERT) &	57.7 &	59.6 &	50.4 &	64.0 &	60.9 &	59.1 &	58.6 \\
\hline
Student-W2V-ISWD	&58.7	& 57.0	& 52.6 &	67.6	& 63.2 &	58.8 &	59.7\\
Student-ATT-ISWD	& 59.6 &	55.9 &	51.0	& \textbf{67.9} &	65.6	& 59.8	& 60.0\\
Student-BERT-ISWD &	59.1 &	\textbf{59.0} &	\textbf{53.9} &	65.8	& \textbf{66.1} &	\textbf{61.0} &	\textbf{60.8}\\
\hline 
\end{tabular}}
\caption{Micro-averaged F1 reported for 9-class EDU-level aspect detection in product reviews.}
\label{tab:aspect-extraction-results-oposum-detailed}
\end{table*}

\end{document}